\documentclass[runningheads]{llncs}
\usepackage[T1]{fontenc}
\usepackage{graphicx,verbatim}
\usepackage{multirow}
\usepackage{booktabs}
\usepackage{amsmath} 
\usepackage{hyperref}
\begin{document}
\title{A Novel Streamline-based diffusion MRI Tractography Registration Method with Probabilistic Keypoint Detection}
\titlerunning{Streamline-based dMRI Registration with Keypoint Detection}

\author{Junyi Wang\inst{1} \and
Mubai Du\inst{1} \and
Ye Wu\inst{2} \and
Yijie Li\inst{1} \and
William M. Wells III\inst{3} \and
Lauren J. O'Donnell\inst{3} \and
Fan Zhang\inst{1}}
%index{Wang, Junyi}
%index{Du, Mubai}
%index{Wu, Ye}
%index{Li, Yijie}
%index{Wells, William}
%index{O'Donnell, Lauren}
%index{Zhang, Fan}

\authorrunning{Junyi Wang et al.}

\institute{
University of Electronic Science and Technology of China, Chengdu, China \and
Nanjing University of Science and Technology, Nanjing, China \and
Harvard Medical School and Brigham and Women’s Hospital, Boston, USA\\
\email{fan.zhang@uestc.edu.cn}
}

\maketitle              % typeset the header of the contribution
\begin{abstract}
Registration of diffusion MRI tractography is an essential step for analyzing group similarities and variations in the brain’s white matter (WM). Streamline-based registration approaches can leverage the 3D geometric information of fiber pathways to enable spatial alignment after registration. Existing methods usually rely on the optimization of the spatial distances to identify the optimal transformation. However, such methods overlook point connectivity patterns within the streamline itself, limiting their ability to identify anatomical correspondences across tractography datasets. In this work, we propose a novel unsupervised approach using deep learning to perform streamline-based dMRI tractography registration. The overall idea is to identify corresponding keypoint pairs across subjects for spatial alignment of tractography datasets. We model tractography as point clouds to leverage the graph connectivity along streamlines. We propose a novel keypoint detection method for streamlines, framed as a probabilistic classification task to identify anatomically consistent correspondences across unstructured streamline sets. In the experiments, we compare several existing methods and show highly effective and efficient tractography registration performance.

% Authors must provide keywords and are not allowed to remove this Keyword section.

\end{abstract}
\section{Introduction}
Diffusion MRI (dMRI) tractography is an advanced tool for in vivo mapping of the brain’s white matter (WM) fiber tracts \cite{Basser2000-ss}. It enables the estimation of the 3D anatomical trajectories of fiber pathways, referred to as the \textit{streamlines}. Registration of tractography streamlines is an essential step in applications such as between-population tract quantification \cite{Chamberland2021-ab,Chandio2020-fb} and fiber tract atlas construction \cite{Zhang2018-tk}. Currently, tractography registration is most often performed by applying transformations derived from registration using volumetric images such as scalar-valued fractional anisotropy (FA) \cite{Moulton2018-cv,Smith2006-wq}, and diffusion models with fiber orientation information \cite{Zhang2022-ko,Zhang2006-gw}. On the other hand, many registration methods are designed to align tractography streamlines directly \cite{Garyfallidis2015-lw,ODonnell2012-fv,Olivetti2016-hg}, which aligns with the scope of our study. Compared to the volumetric registration, streamline-based methods can leverage the 3D geometric information of fiber pathways for improved alignment of fiber tracts. Therefore, streamline-based registration methods hold advantages in the applications with the eventual goal of modeling and analysis of WM fiber tracts. 

Currently, existing streamline-based registration methods employ traditional formulations that solve optimization problems iteratively to align tractography datasets via an energy function \cite{Sotiras2013-ac}. Registration is performed by minimizing the geometric dissimilarity (e.g., Euclidean distance) between the sets of streamlines to spatially align the corresponding fiber tract structures in the brain. However, this process is often computationally intensive due to the need to calculate pairwise streamline similarities. Additionally, steamline misalignment may occur because different anatomical tracts can exhibit similar geometric shapes. Advances in deep learning have shown not only significantly improved accuracy \cite{Fu2020-fb,Haskins2020-jc} but also improved computational efficiency in medical image registration. In related work, recent studies have successfully applied deep learning for improved volumetric dMRI registration \cite{Grigorescu2020-jl,Zhang2022-ko}. However, there are yet no deep learning methods for streamline-based dMRI tractography registration.

One major challenge in performing streamline registration is how to establish the correspondence between streamlines across subjects. In tractography, streamline points are unstructured and represented in Euclidean space, with each point defined by its spatial coordinates (e.g., Right-Anterior-Superior, RAS). This unstructured representation results in the absence of a direct correspondence between tract structures across different subjects. Instead of directly comparing streamlines, a more intuitive approach is to align anatomically equivalent points across subjects within a shared space. This idea is inspired by the recent volume-based registration method, KeyMorph \cite{Wang2023-oe}, which introduces a framework for detecting corresponding keypoints, from which the transformation for registration is then computed. Therefore, we propose to divide the streamline registration process into two sequential steps: (1) detecting corresponding points from moving and fixed tractography, and (2) computing the optimal spatial transformation based on these correspondences. 

In light of the above, we propose a novel streamline-based registration method for dMRI tractography data with a newly proposed streamline probabilistic keypoint detection framework. Our contributions are as follows. First, our method models tractography as a multi-graph structure consisting of interconnected streamlines and thus can leverage the graph connectivity along streamlines to capture structural relationships among the points. To the best of our knowledge, this is the first streamline-based deep learning dMRI tractography registration approach. Second, we introduce a novel keypoint detection strategy formulated as a probabilistic classification task, which enables the identification of anatomically consistent correspondences across inherently unstructured streamline sets. Third, our method is an end-to-end network capable of detecting robust corresponding keypoints across subjects and computing optimal transformations within a unified framework.
\begin{figure}[!t]
    \centering
    \includegraphics[width=\linewidth]{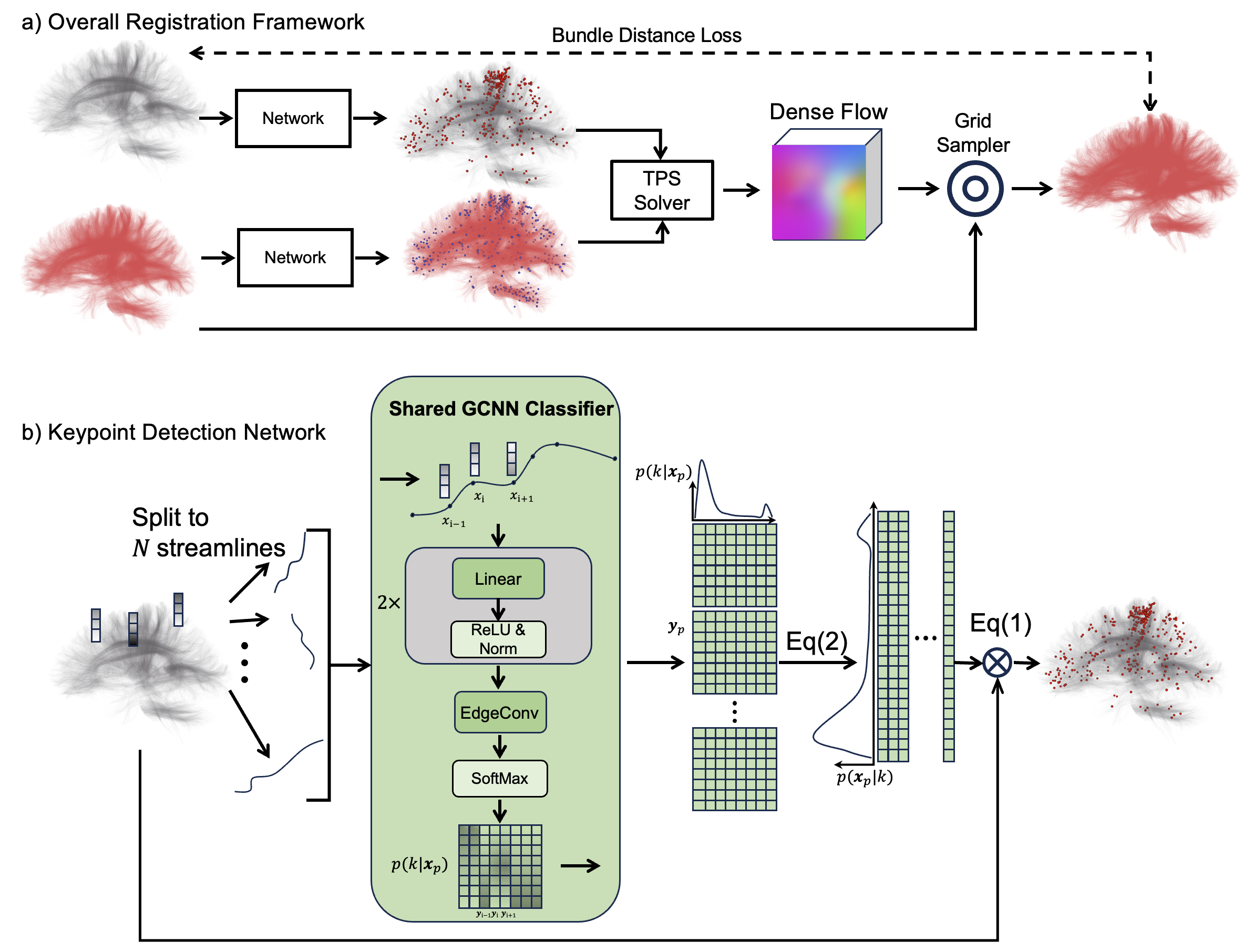}
    \caption{Method Overview. The network consists of two main modules: (1) a keypoint detection network to detect the corresponding keypoint pairs in moving and fixed tractography streamlines, and (2) a thin plate spline (TPS) solver \cite{Bookstein1989-ml,Rohr2001-ha} to predict the transformation based on the keypoint pairs.}
    \label{fig:1}
\end{figure}
\section{Methods}
\subsection{Overall Architecture}
The goal of our method (Fig. \ref{fig:1}a) is to register tractography streamlines between a pair of input subjects. It consists of two main modules: (1) a keypoint detection network to detect the corresponding keypoint pairs in moving and fixed tractography streamlines, and (2) a thin plate spline (TPS) solver \cite{Bookstein1989-ml,Rohr2001-ha} to predict the transformation based on the keypoint pairs. The two modules interact in a feedback loop: the TPS solver’s optimal transformation guides the keypoint detector, refining inter-subject keypoint correspondence, while improved keypoint correspondence enhances the TPS solver’s spatial transformation accuracy.
\subsection{Keypoint Detection Network}
Our keypoint detection network is guided by the hypothesis that anatomically consistent features exist across subjects, differing primarily in spatial position due to individual anatomical variability. The network identifies keypoint locations by modeling their streamline point connectivity relative to the neighboring streamlines rather than relying solely on their spatial coordinates.

\subsubsection{Probabilistic Keypoint Detection.} Consider that an input tractography dataset is composed of $N$ streamlines, each streamline contains $P$ points, and the spatial coordinate of each point is $x_p$. Our goal is to estimate the spatial coordinates of a total of K keypoints. To do so, we model that the spatial coordinate of the \textit{k-th} keypoint $x_k$ follows a distribution $p(x|k)$ that can be estimated based on the distribution of the coordinates of all input points. Specifically, we approximate $p(x|k)$ using $p(x_p|k)$ by discretizing it over all input points that are assumed to follow a uniform distribution. Then, the expected location of the keypoint $x_k$ can be computed as:

\begin{align}
    E\left(x_k\right):=\sum x_p \cdot p(x_p|k)
\end{align}
Here, because $p(x_p|k)$ represents a probability distribution, so as to $p(x_p|k)>0$ and $\sum_{x_p} p(x_p|k)=1$. In this case, the detected keypoints lie within the convex hull formed by all input points, ensuring that all identified keypoints are located inside the brain and not in areas outside the brain regions.
To compute $p(x_p|k)$, we apply Bayes’ theorem \cite{Hirose2021-xt}, as follows:
\begin{align}
    p\left(x_p\middle|k\right)=\frac{p\left(k\middle|x_p\right)}{\sum_{x_p}{p\left(k\middle|x_p\right)}}p\left(x_p\right)
\end{align}
Here, because each point follows a uniform distribution over the entire tractography space, the prior $p(x_p)$ becomes constant across all observations. Consequently, when computing $E(x_k)$, the uniform prior $p(x_p)$ cancels out as a normalization factor. Then, the key to resolving Eq. (2) is computing $p(x_p|k)$. In our study, we propose to use a learnable classifier that computes the probability of each point $x_p$ belonging to the \textit{k-th} keypoint $x_p$, as described below. 

\subsubsection{GCNN keypoint classifier.} To compute the probability $p(k|x_p)$, we train a Graph Convolutional Neural Network (GCNN) classifier \cite{Chen2025-qb,Wang2019-dy} (as shown in Fig. \ref{fig:1}b). We first construct a multigraph based on the streamlines, where each node stores the 3D coordinates of a streamline point, and edges represent the sequential connections along each streamline. Each streamline point \textit{p} first undergoes a feature extraction block, consisting of a linear layer, an activation function, and a normalization to embed \textit{p} into a hidden feature space. Next, we apply edge convolution to the points along each streamline to capture the connectivity patterns within the streamline itself, while simultaneously addressing the disconnectivity between different streamlines. The final layer uses a generalized SoftMax  \cite{Agarwala2020-ag,Barroso-Laguna2023-mq}, which adds a temperature parameter $t$ to standard SoftMax to control output sharpness and enable adjustable confidence calibration. This guarantees that the probabilistic outputs satisfy that $p(k|x_p)>0$ and $\sum_k p(k|x_p)=1$.

\subsection{TPS Solver for Transformation Estimation}

We denote the keypoints detected on moving tractography as $P_k(r, a, s)$, and those on fixed tractography as $Q_k(r, a, s)$. $P_k$ and $Q_k$ are classified into the same anatomical category \textit{k} by the GCNN keypoint classifier, which establishes them as a matched pair. These corresponding keypoint pairs enable us to perform a closed-form solution of optimal transformation: thin plate spline (TPS) \cite{Rohr2001-ha}, which utilizes the correspondence between keypoint pairs to estimate the transformation between two spaces with controllable distortion \cite{Zhao2022-pt}. Specifically. The transformation from the moving space to the fixed space can be computed as
\begin{align}
    \mathcal{T}\left(x_p\right)=A\left[\begin{array}{c}x_p\\1\\\end{array}\right]+\sum_{i=1}^{N}{W_i U\left(\left|P_k-x_p\right|_2\right)}
\end{align}
where $A$ and $W$ are parameters to be solved based on the input keypoints, $U(\cdot)$ is the kernel function ($U(r)=r^2\ln (r)$) \cite{Rohr2001-ha} and $|\cdot|_2$ is the $L^2$ norm. In our work, we utilize the widely used TPS solver proposed in \cite{Rohr2001-ha,Wang2023-oe} to calculate the parameters A and W, as follows:
\begin{align}
    \left[\begin{array}{c}A\\ W\end{array}\right] := \left[\begin{array}{cc}K+\lambda I  & KP\\
                                                                         KP^T & O\end{array}\right]^{-1} \left[\begin{array}{c} KQ^T\\ O\end{array}\right]
\end{align}
where $K_{i j} = U( |P_i - Q_j|_2)$, $O$ is the zero matrix, $\lambda$ is a hyperparameter that controls regularization strength.
\subsection{Loss Function}
We design a loss function to guide the GCNN classifier to find the robust keypoint in both moving tractography $M$ and the fixed $F$. After applying the transformation on moving tractography $M$, we minimize the average minimum distance across all pairs of streamlines from $\mathcal{T}(M)$ and $F$, denoted as follows:
\begin{align}
    Loss\ =\frac{1}{N}\sum\min_{M\in\mathcal{M}}{L_{2,1}\left(\mathcal{T}\left(M\right),F\right)} + \frac{1}{N}\sum\min_{F\in\mathcal{F}}{L_{2,1}\left(\mathcal{T}\left(M\right),F\right)}
\end{align}
where $\mathcal{M}$ and $\mathcal{F}$ represent the sets of tractography streamlines in the moving and fixed datasets, respectively. We adopt the widely used $L_{2,1}$ metric to quantify the distance between the moved tractography $M$ and the fixed $F$ \cite{St-Onge2022-mx}.
\begin{align}
    L_{2,1}\left(\mathcal{T}\left(M\right),F\right)=\frac{1}{P}\sum_{i=1}^{P}\left(\sum_{j=1}^{3}\left|\mathcal{T}\left(m_{ij}\right) - f_{ij}\right|^2\right)^\frac{1}{2}
\end{align}
Here, $m_{i}$ and $f_{i}$ are the coordinates of the \textit{i-th} point on streamline  $M$ and $F$.

\subsubsection{Patch-based Loss.} Training on entire streamline sets is computationally expensive and memory-intensive, as distance calculations scale with the quadratic complexity of pairwise comparisons. To mitigate this, we compute the loss on randomly sampled streamline patches from moved and fixed tractography, reducing the computational burden and memory consumption.

\subsection{Inference}
For inference to register a pair of testing subjects, each input tractography dataset goes through the keypoint detection network (Sec. 2.2). From the detected keypoints of the two subjects, transformation is then estimated using the TPS Solver (Sec. 2.3). Finally, all points on streamlines are aligned into the fixed tractography space by warping their coordinates using the estimated transformation based on Eq (3).

\subsection{Implementation}
Our method is implemented using Pytorch 2.5 \cite{Paszke2019-wx} and executed on a Linux server equipped with NVIDIA 3090 GPUs. We set a total of 512 keypoints to be detected and train a GCNN classifier using the Adam optimizer (initial learning rate: $10^{-3}$, decayed by 0.5 every 10 epochs over 1,000 epochs). Each iteration processes 4 patches of 2,200 streamlines (15 equidistant points per streamline). The generalized SoftMax uses temperature $t = 0.6$, while the TPS solver’s $\lambda$ is sampled log-uniformly during training \cite{Wang2023-oe}. For the inference process, we detect the keypoints on a subset consisting of 30,000 streamlines and then apply the transformation computed by a TPS solver with a fixed $\lambda=0.5$ on whole brain tractography. The implementation of our method will be  available at \href{https://github.com/eiroW/TractoMorph}{\texttt{https://github.com/eiroW/TractoMorph}}

\section{Experiments and Results}
\subsection{Experimental Datasets}
We use the tractography data provided in TractSeg \cite{Wasserthal2018-cg} (https://doi.org/10.5281/ zenodo.1088277), derived from dMRI data from 105 subjects in the Human Connectome Project (HCP) Young Adult \cite{Van_Essen2013-ae}. For each subject, the provided whole brain tractography data contains about 1.6 million streamlines. We choose to use this data because each streamline is associated with an anatomical label for experimental evaluation (Sec. 3.2). In our experiments, we allocate 90 subjects for training, 5 subjects for validation, and 10 subjects for testing.

\subsection{Experimental Design}
\subsubsection{Comparison to SOTA Methods.} We evaluate our method against both volume-based and streamline-based registration methods. To compare with unsupervised volume-based methods include SyN \cite{Avants2008-es} in ANTs and SynthMorph \cite{Hoffmann2022-ds} in FreeSurfer, we adjust the registration pipeline, taking b0 images as input to compute spatial transformations, followed by warping the moving tractography data using 3D Slicer \cite{Fedorov2012-ki,Zhang2020-pf}. Streamline-based methods include Streamline-based Linear Registration (SLR) \cite{Chandio2020-fb} from DiPy and a nonlinear registration method in White Matter Analysis (WMA) \cite{Basser2000-ss}. These methods directly process whole-brain tractography datasets to align streamlines.

\subsubsection{Ablation Study.} The key of our method is to leverage the corresponding keypoints detected using our Keypoint Detection Network. To assess its performance, we replace it with a nearest-neighbor (NN) algorithm that randomly samples keypoints from the input tractography datasets. The correspondence between these keypoints is computed based on a Euclidean distance-based NN. The resulting correspondences are fed into the TPS solver to estimate a spatial transformation which is applied to align the entire moving tractography.

\subsubsection{Evaluation Metrics.} To quantify registration quality, we assess whether the corresponding anatomical fiber bundles are aligned after registration. This take advantage of  the predefined anatomical labels of each steamline provided in the experimental tractography data. For each fiber bundle, two metrics are computed: weighted Dice Score (wDice) \cite{Crum2006-xq} and the Average Bundle Distance (ABD) \cite{Garyfallidis2015-lw} . The wDice extends the standard Dice metric by incorporating tract intensity, weighting each class by the inverse of its tract intensity to balance their contributions to the overall similarity. The ABD is defined as the average of the minimum Mean Direct-Flip (MDF) distances\cite{Garyfallidis2012-ql} between each streamline in the moving set and all streamlines in the fixed set, and vice versa.
\begin{figure}[!t]
    \centering
    \includegraphics[width=1\linewidth]{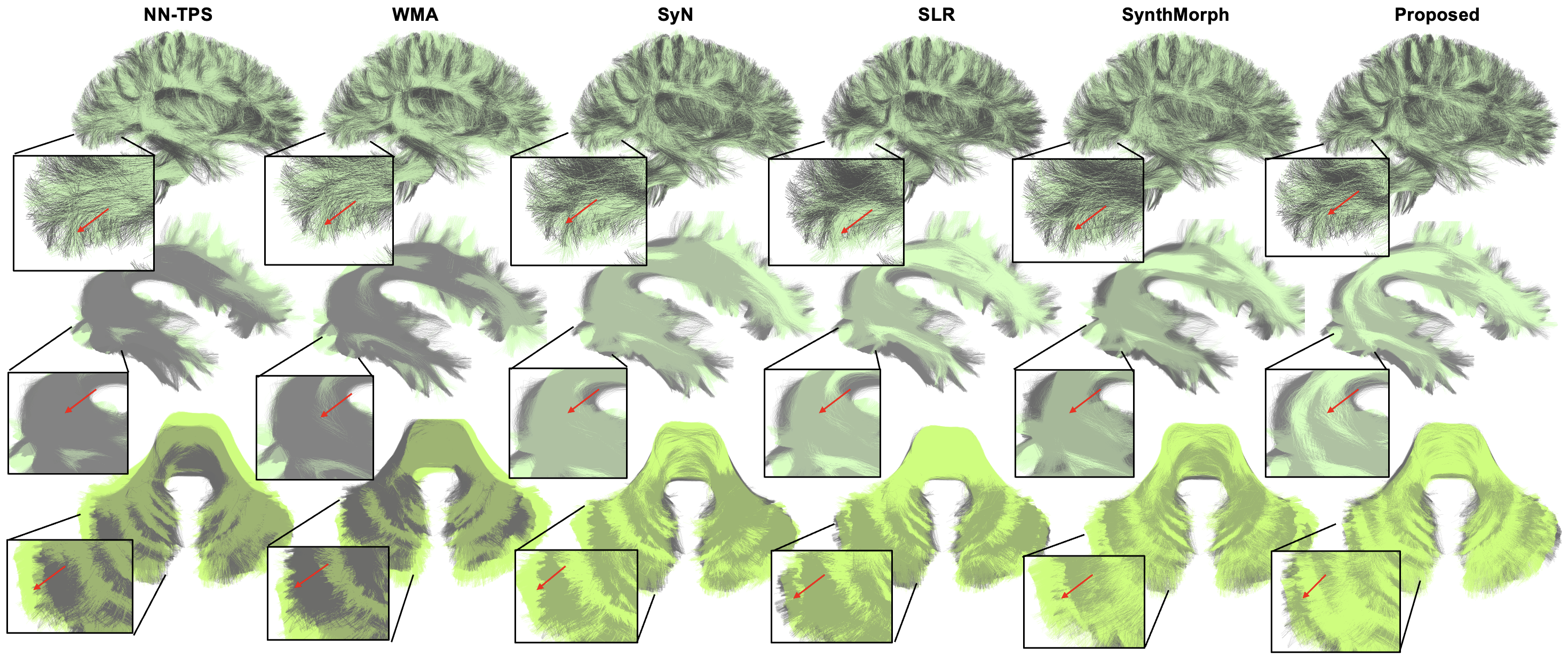}
    \caption{Visualization of registration performance across different methods. Our method generally results in a better spatial overlap between two tracts than other methods, yielding a more merged and parallel alignment, depicted as a more merged color.}
    \label{fig:2}
\end{figure}

\subsection{Results}
\subsubsection{Quantitative Results.} Table \ref{tab1} shows the results of the quantitative comparison between all the volume- and streamline-based methods. Our method obtains the lowest ABD with all pairwise t-tests showing FDR-adjusted p < 0.001 compared to other methods, indicating superior spatial alignment. Regarding wDice, our method overperforms all streamline-based methods, and also the voxel-wise ANTs(FDR-adjusted p < 0.001), except for SynthMorph.  One potential reason is that SynthMorph operates on dense volumes, so it can better find volumetric overlap. In particular, our method registers all 1.6 million moving streamlines to the fixed space in approximately 15 seconds on average, outperforming all other registration methods in terms of speed.

\begin{table}[!t]
\caption{Comparison of wDice and ABD across different methods.}\label{tab1}
\centering
\begin{tabular}{llcccc}
\hline
\multicolumn{2}{l}{\multirow{2}{*}{Methods}} & \multicolumn{2}{c}{ABD} & \multicolumn{2}{c}{wDice} \\
\multicolumn{2}{l}{}                        & mean(mm)& \hspace{5pt}{std}\hspace{5pt}      & \hspace{5pt}{mean}\hspace{5pt}        & \hspace{5pt}{std}\hspace{5pt}\\
\hline
% \multicolumn{2}{l}{Volume-based}             &              &          &             &             \\
\multirow{2}{*}{\textbf{Volume-based}}& SyN               & 3.794        & 1.054    & 70.3\%      & 15.6\%      \\
                         & SynthMorph        & 3.468        & 0.947    & \textbf{77.9}\%      & 14.0\%      \\
                        \hline
% \multicolumn{2}{l}{Streamline-based}         &              &          &             &             \\
\multirow{4}{*}{\textbf{Streamline-based}}& WMA               & 3.899        & 0.928    & 65.1\%      & 15.5\%      \\
                         & SLR               & 3.557        & 0.738    & 70.7\%      & 12.9\%      \\
                         & NN+TPS            & 4.530        & 1.58     & 58.2\%      & 19.2\%      \\
                         & Proposed          & \textbf{3.338}        & \textbf{0.711}    & 74.8\%      & \textbf{12.5}\%     \\
\hline
\end{tabular}
\end{table}

\subsubsection{Visual Results.} In Fig \ref{fig:2}, we can observe that our method generally results in a better spatial overlap between two tracts than other methods, yielding a more merged and parallel alignment (depicted as a more merged color). Additionally, in the black-circled regions, our method achieves better local direction alignment compared to the other methods. %In particular, the network detects significant misalignments in the circled areas. For MCPs of this subject pair, the displacement captured is larger in the middle of the tract compared to the endpoint regions.

\begin{figure}[!t]
    \centering
    \includegraphics[width=0.95\linewidth]{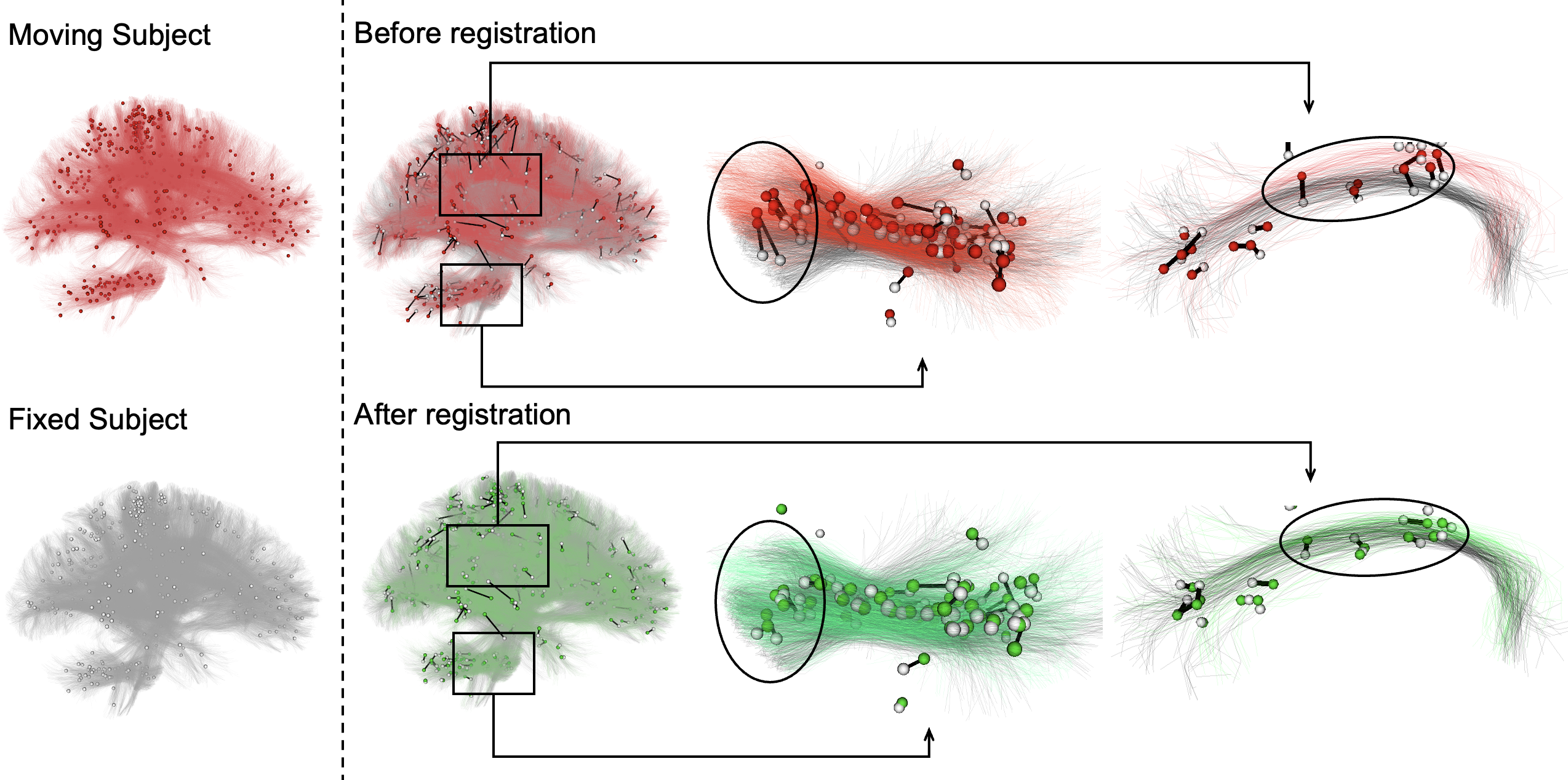}
    \caption{Left: Keypoints detected in moving and fixed tractography. Right: Keypoint correspondence (whole brain, AF and MCP) before and after registration. The correspondences between subjects and the displacement is highlight in black. }
    \label{fig:3}
\end{figure}

Furthermore, we visualize the detected keypoints between moving and fixed subjects before and after registration (Fig. \ref{fig:3}). The results show that the network successfully identifies correspondences between subjects and accurately captures the displacements across different parts of the tracts (arcuate fasciculus, AF, and middle cerebellar peduncle, MCP). These identified keypoints as robust anchors for estimating the nonlinear spatial transformation, enabling precise alignment of entire streamline sets.

\section{Conclusion}
This work proposes a novel streamline-based registration method for dMRI tractography data. Our method is an end-to-end unsupervised learning framework that is capable of detecting robust corresponding keypoints across subjects and computing optimal transformations within a unified framework. Our results show highly effective and efficient registration performance and demonstrate the benefit of using deep learning for tractography registration tasks. 

\subsubsection{Acknowledgments} This work is supported by the National Key R\&D Program of China (No. 2023YFE0118600) and the National Natural Science Foundation of China (No. 62371107).

\subsubsection{Disclosure of Interest} The authors have no competing interests to declare that are relevant to the content of this article.
% \subsubsection{Disclosure of Interests} The authors have no competing interests to declare that are relevant to the content of this article.
\bibliographystyle{splncs04}
\bibliography{Paper-3289}

\begin{thebibliography}{10}
\providecommand{\url}[1]{\texttt{#1}}
\providecommand{\urlprefix}{URL }
\providecommand{\doi}[1]{https://doi.org/#1}

\bibitem{Agarwala2020-ag}
Agarwala, A., Pennington, J., Dauphin, Y., Schoenholz, S.: Temperature check: theory and practice for training models with softmax-cross-entropy losses. arXiv [cs.LG]  (Oct 2020)

\bibitem{Avants2008-es}
Avants, B.B., Epstein, C.L., Grossman, M., Gee, J.C.: Symmetric diffeomorphic image registration with cross-correlation: evaluating automated labeling of elderly and neurodegenerative brain. Med. Image Anal.  \textbf{12}(1),  26--41 (Feb 2008)

\bibitem{Barroso-Laguna2023-mq}
Barroso-Laguna, A., Mikolajczyk, K.: Key.net: Keypoint detection by handcrafted and learned {CNN} filters revisited. IEEE Trans. Pattern Anal. Mach. Intell.  \textbf{45}(1),  698--711 (Jan 2023)

\bibitem{Basser2000-ss}
Basser, P.J., Pajevic, S., Pierpaoli, C., Duda, J., Aldroubi, A.: In vivo fiber tractography using {DT}-{MRI} data. Magn. Reson. Med.  \textbf{44}(4),  625--632 (Oct 2000)

\bibitem{Bookstein1989-ml}
Bookstein, F.L.: Principal warps: thin-plate splines and the decomposition of deformations. IEEE Trans. Pattern Anal. Mach. Intell.  \textbf{11}(6),  567--585 (Jun 1989)

\bibitem{Chamberland2021-ab}
Chamberland, M., Genc, S., Tax, C.M.W., etc.: Detecting microstructural deviations in individuals with deep diffusion {MRI} tractometry. Nat Comput Sci  \textbf{1},  598--606 (Sep 2021)

\bibitem{Chandio2020-fb}
Chandio, B.Q., Risacher, S.L., etc.: Bundle analytics, a computational framework for investigating the shapes and profiles of brain pathways across populations. Sci. Rep.  \textbf{10}(1),  17149 (Oct 2020)

\bibitem{Chen2025-qb}
Chen, Y., Zhang, F., Wang, M., etc.: {TractGraphFormer}: Anatomically informed hybrid graph {CNN}-transformer network for interpretable sex and age prediction from diffusion {MRI} tractography. Med. Image Anal.  \textbf{101}(103476) (Jan 2025)

\bibitem{Crum2006-xq}
Crum, W.R., Camara, O., Hill, D.L.G.: Generalized overlap measures for evaluation and validation in medical image analysis. IEEE Trans. Med. Imaging  \textbf{25}(11),  1451--1461 (Nov 2006)

\bibitem{Fedorov2012-ki}
Fedorov, A., Beichel, R., Kalpathy-Cramer, J., etc.: {3D} slicer as an image computing platform for the quantitative imaging network. Magn. Reson. Imaging  \textbf{30}(9),  1323--1341 (Nov 2012)

\bibitem{Fu2020-fb}
Fu, Y., Lei, Y., Wang, T., Curran, W.J., Liu, T., Yang, X.: Deep learning in medical image registration: a review. Phys. Med. Biol.  \textbf{65}(20),  20TR01 (Oct 2020)

\bibitem{Garyfallidis2012-ql}
Garyfallidis, E., Brett, M., Correia, M.M., Williams, G.B., Nimmo-Smith, I.: {QuickBundles}, a method for tractography simplification. Front. Neurosci.  \textbf{6}, ~175 (Dec 2012)

\bibitem{Garyfallidis2015-lw}
Garyfallidis, E., Ocegueda, O., Wassermann, D., Descoteaux, M.: Robust and efficient linear registration of white-matter fascicles in the space of streamlines. Neuroimage  \textbf{117},  124--140 (Aug 2015)

\bibitem{Grigorescu2020-jl}
Grigorescu, I., Uus, A., Christiaens, D., etc.: Diffusion tensor driven image registration: A deep learning approach. In: Biomedical Image Registration. pp. 131--140. Springer International Publishing (2020)

\bibitem{Haskins2020-jc}
Haskins, G., Kruger, U., Yan, P.: Deep learning in medical image registration: a survey. Mach. Vis. Appl.  \textbf{31}(1), ~8 (Jan 2020)

\bibitem{Hirose2021-xt}
Hirose, O.: A bayesian formulation of coherent point drift. IEEE Trans. Pattern Anal. Mach. Intell.  \textbf{43}(7),  2269--2286 (Jul 2021)

\bibitem{Hoffmann2022-ds}
Hoffmann, M., Billot, B., Greve, D.N., Iglesias, J.E., Fischl, B., Dalca, A.V.: {SynthMorph}: Learning contrast-invariant registration without acquired images. IEEE Trans. Med. Imaging  \textbf{41}(3),  543--558 (Mar 2022)

\bibitem{Moulton2018-cv}
Moulton, E., Valabregue, R., etc.: Comparison of spatial normalization strategies of diffusion {MRI} data for studying motor outcome in subacute-chronic and acute stroke. Neuroimage  \textbf{183},  186--199 (Dec 2018)

\bibitem{ODonnell2012-fv}
O'Donnell, L.J., Wells, 3rd, W.M., Golby, A.J., Westin, C.F.: Unbiased groupwise registration of white matter tractography. Med. Image Comput. Comput. Assist. Interv.  \textbf{15}(Pt 3),  123--130 (2012)

\bibitem{Olivetti2016-hg}
Olivetti, E., Sharmin, N., Avesani, P.: Alignment of tractograms as graph matching. Front. Neurosci.  \textbf{10}, ~554 (Dec 2016)

\bibitem{Paszke2019-wx}
Paszke, A., Gross, S., Massa, F., Lerer, A., Bradbury, J., Chanan, G., Killeen, T., Lin, Z., Gimelshein, N., Antiga, L., {Others}: Pytorch: An imperative style, high-performance deep learning library. Adv. Neural Inf. Process. Syst.  \textbf{32} (2019)

\bibitem{Rohr2001-ha}
Rohr, K., Stiehl, H.S., Sprengel, R., Buzug, T.M., Weese, J., Kuhn, M.H.: Landmark-based elastic registration using approximating thin-plate splines. IEEE Trans. Med. Imaging  \textbf{20}(6),  526--534 (Jun 2001)

\bibitem{Smith2006-wq}
Smith, S.M., Jenkinson, M., Johansen-Berg, H., etc.: Tract-based spatial statistics: voxelwise analysis of multi-subject diffusion data. Neuroimage  \textbf{31}(4),  1487--1505 (Jul 2006)

\bibitem{Sotiras2013-ac}
Sotiras, A., Davatzikos, C., Paragios, N.: Deformable medical image registration: A survey. IEEE Trans. Med. Imaging  \textbf{32}(7),  1153--1190 (Jul 2013)

\bibitem{St-Onge2022-mx}
St-Onge, E., Garyfallidis, E., Collins, D.L.: Fast streamline search: An exact technique for diffusion {MRI} tractography. Neuroinformatics  \textbf{20}(4),  1093--1104 (Oct 2022)

\bibitem{Van_Essen2013-ae}
Van~Essen, D.C., Smith, S.M., Barch, D.M., Behrens, T.E.J., Yacoub, E., Ugurbil, K., {WU-Minn HCP Consortium}: The {WU}-minn human connectome project: an overview. Neuroimage  \textbf{80},  62--79 (Oct 2013)

\bibitem{Wang2023-oe}
Wang, A.Q., Yu, E.M., Dalca, A.V., Sabuncu, M.R.: A robust and interpretable deep learning framework for multi-modal registration via keypoints. Med. Image Anal.  \textbf{90}(102962),  102962 (Dec 2023)

\bibitem{Wang2019-dy}
Wang, Y., Sun, Y., Liu, Z., etc.: Dynamic graph {CNN} for learning on point clouds. ACM Trans. Graph.  \textbf{38}(5),  1--12 (Oct 2019)

\bibitem{Wasserthal2018-cg}
Wasserthal, J., Neher, P., Maier-Hein, K.H.: {TractSeg} - fast and accurate white matter tract segmentation. Neuroimage  \textbf{183},  239--253 (Dec 2018)

\bibitem{Zhang2020-pf}
Zhang, F., Noh, T., Juvekar, P., etc.: {SlicerDMRI}: Diffusion {MRI} and tractography research software for brain cancer surgery planning and visualization. JCO Clin Cancer Inform  \textbf{4},  299--309 (Mar 2020)

\bibitem{Zhang2022-ko}
Zhang, F., Wells, W.M., O'Donnell, L.J.: Deep diffusion {MRI} registration ({DDMReg}): A deep learning method for diffusion {MRI} registration. IEEE Trans. Med. Imaging  \textbf{41}(6),  1454--1467 (Jun 2022)

\bibitem{Zhang2018-tk}
Zhang, F., Wu, Y., etc. Rathi, Y., Makris, N., O'Donnell, L.J.: An anatomically curated fiber clustering white matter atlas for consistent white matter tract parcellation across the lifespan. Neuroimage  \textbf{179},  429--447 (Oct 2018)

\bibitem{Zhang2006-gw}
Zhang, H., Yushkevich, P.A., Alexander, D.C., Gee, J.C.: Deformable registration of diffusion tensor {MR} images with explicit orientation optimization. Med. Image Anal.  \textbf{10}(5),  764--785 (Oct 2006)

\bibitem{Zhao2022-pt}
Zhao, J., Zhang, H.: Thin-plate spline motion model for image animation. Proc. IEEE Comput. Soc. Conf. Comput. Vis. Pattern Recognit. pp. 3647--3656 (Mar 2022)

\end{thebibliography}

\end{document}